\documentclass{article}

\usepackage[section]{placeins}
\usepackage{times}
\usepackage{calc}
\usepackage{float}
\usepackage{ragged2e}
\usepackage[numbers,sort]{natbib}
\usepackage{PRIMEarxiv}
\usepackage[super]{nth}
\usepackage[utf8]{inputenc} 
\usepackage[T1]{fontenc}    
\usepackage{hyperref}       
\usepackage{url}            
\usepackage{booktabs}       
\usepackage{amsfonts}       
\usepackage{nicefrac}       
\usepackage{microtype}      
\usepackage{lipsum}
\usepackage{fancyhdr}       
\usepackage{graphicx}       
\usepackage{atbegshi}
\graphicspath{{media/}}     

\title{Real-Time Superficial Vein Imaging System for Observing Abnormalities on Vascular Structures}

\newcommand\blfootnote[1]{%
  \begingroup
  \renewcommand\thefootnote{}\footnote{#1}%
  \addtocounter{footnote}{-1}%
  \endgroup
}
\author{\large
  Ayse Altay, Abdurrahman Gumus*
  \\\large
  \\ Department of Electrical and Electronics Engineering, Izmir Institute of Technology}
 
\begin{document}

\maketitle

\begin{abstract}
Circulatory system abnormalities might be an indicator of diseases or tissue damage. Early detection of vascular abnormalities might have an important role during treatment and also raise the patient’s awareness. Current detection methods for vascular imaging are high-cost, invasive, and mostly radiation-based. In this study, a low-cost and portable microcomputer-based tool has been developed as a near-infrared (NIR) superficial vascular imaging device. The device uses NIR light-emitting diode (LED) light at 850 nm along with other electronic and optical components. It operates as a non-contact and safe infrared (IR) imaging method in real-time. Image and video analysis are carried out using OpenCV (Open-Source Computer Vision), a library of programming functions mainly used in computer vision. Various tests were carried out to optimize the imaging system and set up a suitable external environment. To test the performance of the device, the images taken from three diabetic volunteers, who are expected to have abnormalities in the vascular structure due to the possibility of deformation caused by high glucose levels in the blood, were compared with the images taken from two non-diabetic volunteers. As a result, tortuosity was observed successfully in the superficial vascular structures, where the results need to be interpreted by the medical experts in the field to understand the underlying reasons. Although this study is an engineering study and does not have an intention to diagnose any diseases, the developed system here might assist healthcare personnel in early diagnosis and treatment follow-up for vascular structures and may enable further opportunities. \blfootnote{* Corresponding author: abdurrahmangumus@iyte.edu.tr \\ \\ 
Machine Intelligence Research and Applications Laboratory (MIRALAB) \\Preprint submitted. Under review.}
\end{abstract}


\section{Introduction}
\label{sec:introduction}
Vascular imaging techniques, such as tomography, magnetic resonance imaging, angiogram, and vascular ultrasound, play a critical role in the diagnosis and management of a variety of diseases and conditions in today’s technology. Even though the traditional techniques can give high-resolution images of vascular structures, they have some disadvantages. Tomography needs to use an iodinated contrast agent to enhance the visibility of vascular structure, but ionizing radiation can be harmful for long-term use \cite{Kramer2007}. Also, the contrast agent may cause allergic reaction for some people. Magnetic resonance imaging (MRI) does not use ionizing radiation and contrast agent. However, the technique has high-cost and trained personnel limitations along with requiring a large space for equipment operation \cite{Dale2015}. Magnetic resonance angiography (MRA) and computed tomography angiography (CTA) are some types of angiogram techniques. CTA has a type of radiation that may damage DNA and may cause cancer \cite{van2018contrast}. On the other hand, MRA does not use ionizing radiation and there is no need for contrast agents, but sometimes can induce a feeling of claustrophobia \cite{goyen2000mr}. Ultrasound needs trained staff due to the complex operating structure of the device \cite{Norris2004}. Compared to other imaging techniques, although ultrasound is a low-cost, it is a direct contact method on the body with a small amount of ultrasound gel or cream, which can cause minor skin irritation or allergic reactions for some people. Compared to traditional imaging techniques, near-infrared (NIR) imaging has many advantages due to being low-cost, non-contact, portable, and user-friendly.

The NIR light wavelength is between 700-900 nm in the electromagnetic spectrum \cite{curtin2012file} and cannot be seen by human eyes. NIR light is non-ionizing since it does not have enough energy for ionization which reduces the possibility of tissue damage. In the blood, there are two absorbers of NIR light as oxyhemoglobin (HbO2) and deoxyhemoglobin (Hb). In the absorption spectra for oxy- and deoxy-hemoglobin between the 600 and 1000 nm range, HbO2 absorbs more IR light and transmits more red light than Hb, whereas Hb transmits more IR light and absorbs more red light than HbO2. NIR light illuminates the target area where the light passes through the skin and is absorbed by the hemoglobin. In the NIR region, hemoglobin absorption is higher than in skin and fat tissue, and also, the absorption of oxy- and deoxyhemoglobin peaks have a maximum value of 850 nm. In this way, blood vessels appear darker than other surrounding tissue \cite{Ai2016,D'Alessandro2012} which makes the NIR region advantageous to image the vascular structures. There are some studies using NIR imaging technology focusing on authentication-based applications such as the finger, hand, palm, and wrist vein detection \cite{Lee2011,Huang2017,Lingyu2006}. In addition, NIR light has the potential to safely and effectively treat tissue stressed by mitochondrial dysfunction \cite{wan2020nir,Kobayashi2019,Desmet2006} and stop neurodegeneration, that is, progressive loss of structure or function of neurons, in the photobiomodulation therapy which is performed for the treatment of neurological disorders such as Alzheimer's and Parkinson's. Therefore, NIR light is a good alternative for human applications \cite{Sowa2016,Johnstone2016,Schiffer2009,Barolet2016}. Even though there are studies investigating vascular structure using NIR imaging based in the medical field \cite{Zharov2004,Abdeladl2016}, low-cost devices with real-time imaging and video analysis capability are still highly desired.

\begin{figure}[htbp] \centering
\includegraphics[width=0.65\textwidth]{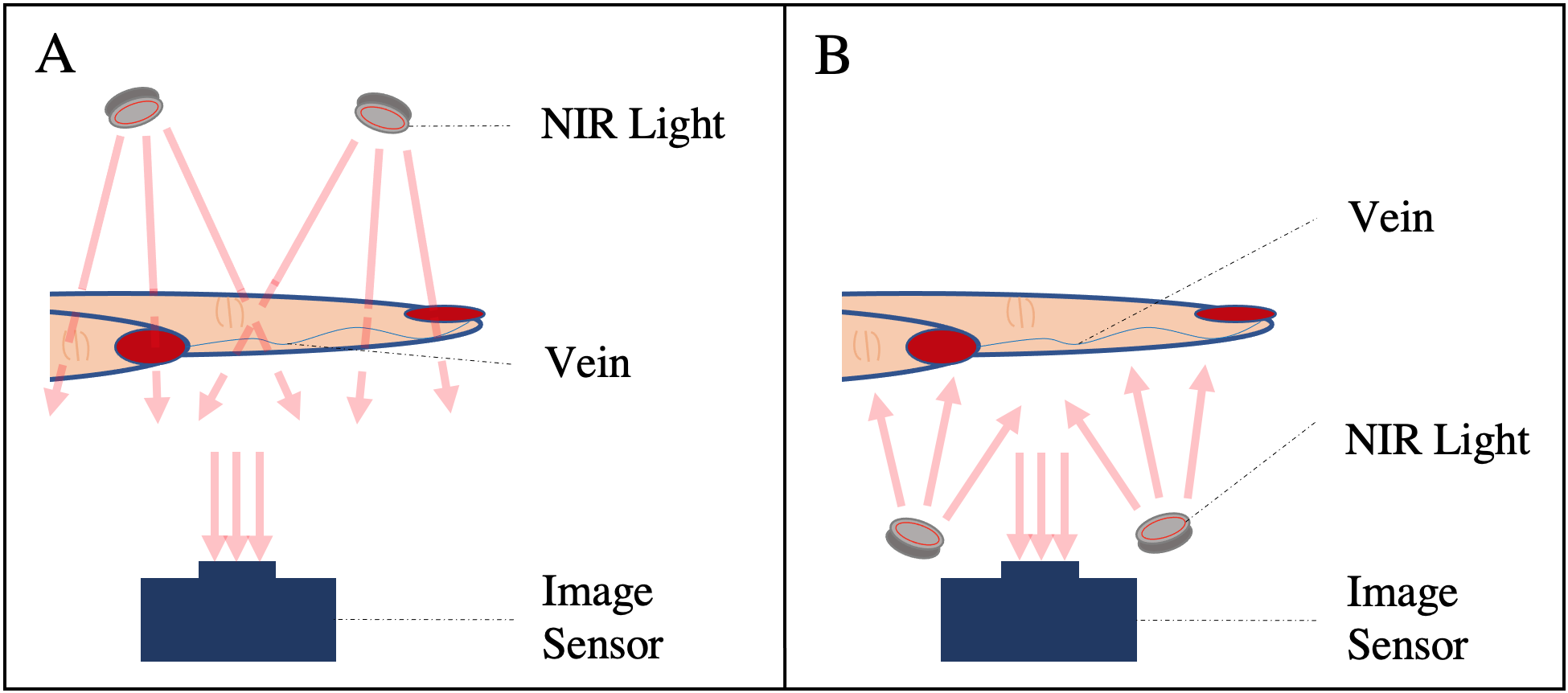}
\caption{NIR Imaging Module. (A) Transmission of light. (B) Reflection of light.}
    \label{fig:fig1}
\end{figure}

There are two main NIR vascular imaging methods that have been used in previous studies, which are the transmission of light (Figure \ref{fig:fig1}A) and the reflection of light (Figure \ref{fig:fig1}B). The reflection technique uses light reflection from a target. Blood vessels appeared dark whereby IR light on images captured with an IR camera. This technique is more appropriate to design portable devices with a small footprint and to image large parts of the human body without the need of transmitting light through. Yildiz $\emph{et al.}$ \cite{yildiz2019development} reported that a low-cost microcomputer based dorsal hand-vein imaging system using reflection method for their system in the study. In the light transmission technique, which is an effective method to obtain a high contrast image in vascular imaging, the NIR light passes through the skin directly without spreading and the light energy is mostly conserved. In this case, most of the light is transmitted to the image sensor. This technique can have high-quality blood vessel images but requires large-size devices due to the location of the sensors and the light sources. Therefore, it is more suitable for imaging the thin parts of the human body. Mela $\emph{et al.}$ \cite{mela2019real} used light transmission method for NIR images. They combined them with the visible spectrum images taken with reflection method in real time to achieve vein localization. In addition, there is a third technique using transmission, called the side lighting method, proposed by Hashimoto $\emph{et al.}$ \cite{Hashimoto2006}. In this method, the target is illuminated from both sides with NIR light and an IR camera is placed at an angle of 90 degrees to the lights. The technique is suitable for imaging the thin parts of the human body like the light transmission method.

The circulatory system supplies blood circulation to the cells and organs in the body to provide mainly oxygen and nutrients while removing the waste. Any abnormal condition in the vascular system may result in serious health consequences such as vascular diseases and disabilities. Many vascular diseases are caused by the fact that the blood vessels do not work effectively to supply blood to the organs due to enlargement, narrowing, blockage, or damage of vessels in the body. In addition, unhealthy diet, smoking, genetics, advanced age, insufficiency of physical activity, and complications resulting from diseases such as diabetes, hypertension, and high cholesterol may also cause these vascular diseases. Early diagnosis plays a critical role in their treatment.  Therefore, it is important for people who are at risk to check their vascular health regularly, safely, and easily. Imaging is needed not only in early diagnosis but also in treatments such as sclerotherapy, which is used in the treatment of varicose and spider veins. During the procedure, a solution is injected into the blood vessels. The treatment process can require real-time monitoring in order to accurately identify the superficial veins. A real-time, portable, user-friendly, non-radiation, and non-invasive device will provide convenience to specialists for superficial vein monitoring. 

Diabetes mellitus is one of the most common chronic and metabolic diseases globally. Insulin insufficiency or insulin resistance of the cells can cause increased sugar in the blood; as a result, organs can get damaged. In 2020, an estimated 422 million adults had diabetes, and 1.5 million deaths each year are directly related to diabetes. Diabetes has different types such as Type 1 and Type 2. More than 95\% of people with diabetes have Type 2 diabetes. There are two interrelated problems in type 2 diabetes: The pancreas that regulates the transport of sugar in the body to the cells does not produce enough insulin and the cells respond poorly to insulin \cite{world2020guideline}. Almér $\emph{et al.}$ \cite{almer1975plasminogen} reported that the fibrinolytic response, a process that prevents blood clots from growing and becoming problematic in diabetic patients, is weak against venous occlusion. Complications of diabetes can slow down the healing of wounds, and also cause nerve or tissue damage, and in some cases, amputation may occur. Diabetic foot ulcer, caused by the long-term chronic condition of diabetes, is the most common reason for foot amputation in the world. Additionally, diabetic foot treatments and cares are costly and require frequent hospital visits \cite{Raspovic2014}. Regular vascular evaluation is important to detect patients with a high risk of vascular abnormalities for early diagnosis \cite{Beckert2006,Marshall2006,Apelqvist2000,schaper2012diagnosis}. For example, detection of reduced or impaired blood circulation in a diabetic foot patient would be necessary to prevent an amputation that may occur in the future. 

\begin{figure}[htbp] \centering
\includegraphics[width=0.65\textwidth]{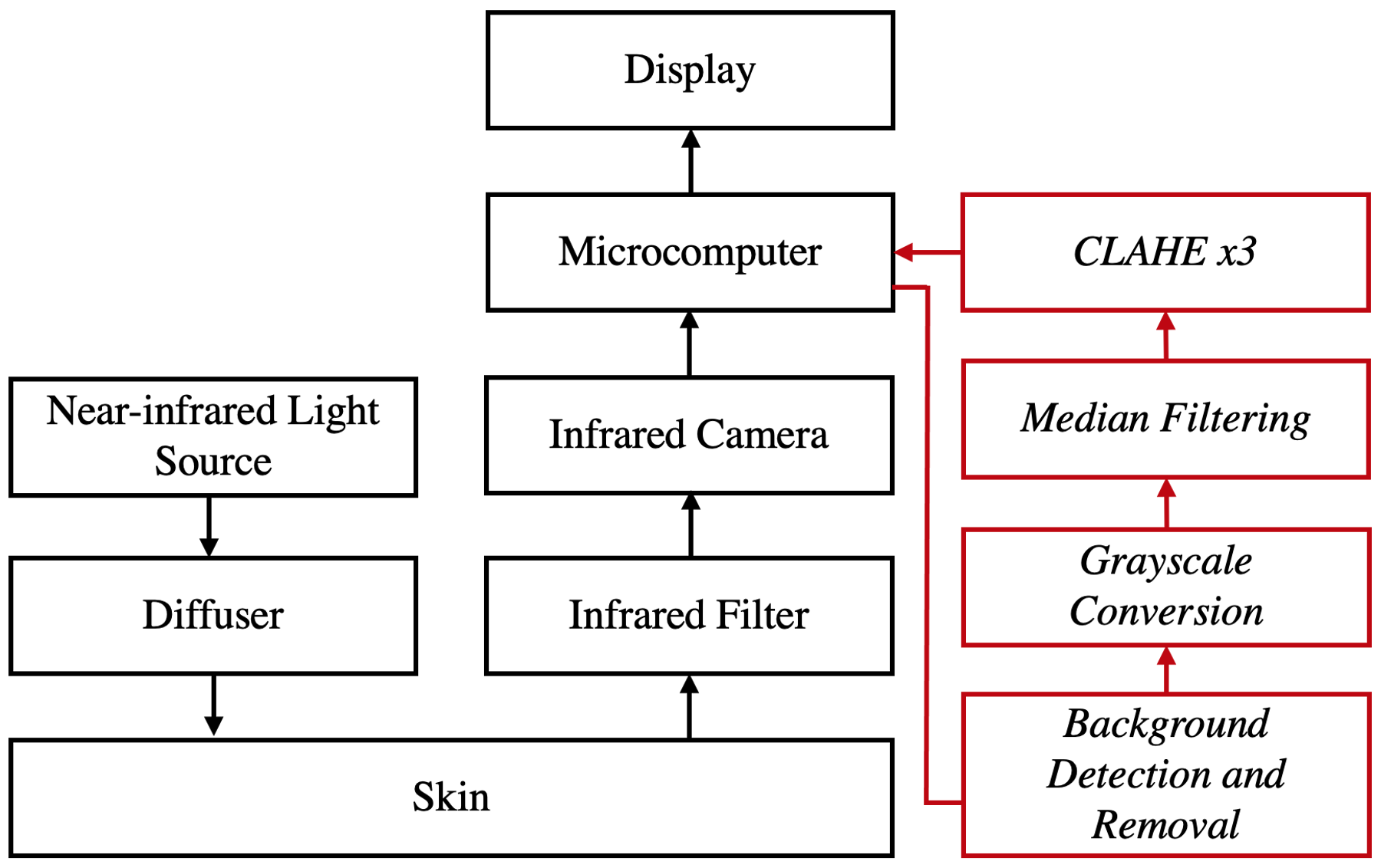}
\caption{Block diagram of the system containing hardware and software components.}
    \label{fig:fig2}
\end{figure}

In this study, a low-cost and portable device was designed, built, and used to obtain images of vascular structures using NIR light. Our goal is to contribute to the field in a way that helps experts with the early diagnosis of vascular diseases or by monitoring the changes during a treatment process in a safe and easy manner with the real-time imaging system. It has been reported that NIR light penetrates to the depths of 2.6 and 15 mm under the skin \cite{Ai2016,Zharov2004,Seker2017,Cuper2013,Kim2017,Miyake2006}. The depth of the hand and foot superficial vein range from 2 to 4.1 mm \cite{Brandt2016}. Therefore, we anticipate that the developed device is capable of taking images at a depth of approximately 2.8 - 4 mm under the skin. Our method uses similar approaches to those used in NIR-based biometric identification systems to produce high-quality images. The block diagram in Figure \ref{fig:fig2} summarizes the hardware and software used in this work. The performance of the built device has been investigated by capturing some foot and leg images from diabetic and non-diabetic volunteers. 

The rest of the paper is organized as follows: In the first section, hardware and software implementation details are reviewed, and proposed device fabrication is explained. Then, the experimental results are presented and discussed. Concluding remarks and future perspectives are given in the last section.

\section{Materials and Methods}

\subsection{Hardware Implementation}

The design of the hardware system consists of two NIR lights, a NIR camera module, special IR filter in the range of 700–900 nm at average wavelength, LED light compatible diffusers and a microcomputer. NIR lights consist of 1W high power IR illuminator modules with 850 nm wavelength, adjustable resistance, light dependent resistance (LDR) and lens \cite{Taoyuan}. They also have adjustable screw to change the brightness of the daylight and night light sensitivity threshold. Raspberry Pi NoIR Camera Module is an 8-megapixel Sony IMX219 image sensor, which has a fixed focus lens without an IR filter. The camera is capable of 3280 x 2464 pixels static display and also supports 1080p30, 720p60 and 640x480p60 / 90 video \cite{Raspberry}. Supported by Nvidia Jetson Nano, this camera connects to the small socket on top of the card via a ribbon cable and uses a special Camera Serial Interface (CSI) specifically designed to interface with cameras. The size and small weight of the camera provide convenience for many applications. IR filters are special filters designed to pass light in the NIR wavelength range of 700–900 nm on average. IR filter was used to eliminate ambient light to be able to acquire images during daylight. The IR filter used in our study transmits 95\% of the light in the range of 760 nm – 860 nm \cite{Hoya}. Diffusers are optical components that are used to improve the smoothing of the illumination intensity, diffuse the light, and provide a better visualization \cite{Yongnuo}. The Nvidia Jetson Nano Developer Kit (70 x 45 mm) is one of the smallest microcomputers among Jetson devices. Since Jetson Nano has low power consumption (5-10 watts) and high processing power up to 472 GFLOPS, it can quickly run image processing, and machine learning algorithms \cite{Nvidia}. Jetson Nano Developer Kit consists of a microSD card slot, 40-pin expansion header, micro-USB port, gigabit ethernet port, 4 USB ports, HDMI output port, display port connector, DC barrel jack for 5V power input, and CSI camera connectors. 

\subsection{Device Design for 3D Modeling}

\begin{figure}[htbp] \centering
\includegraphics[width=0.70\textwidth]{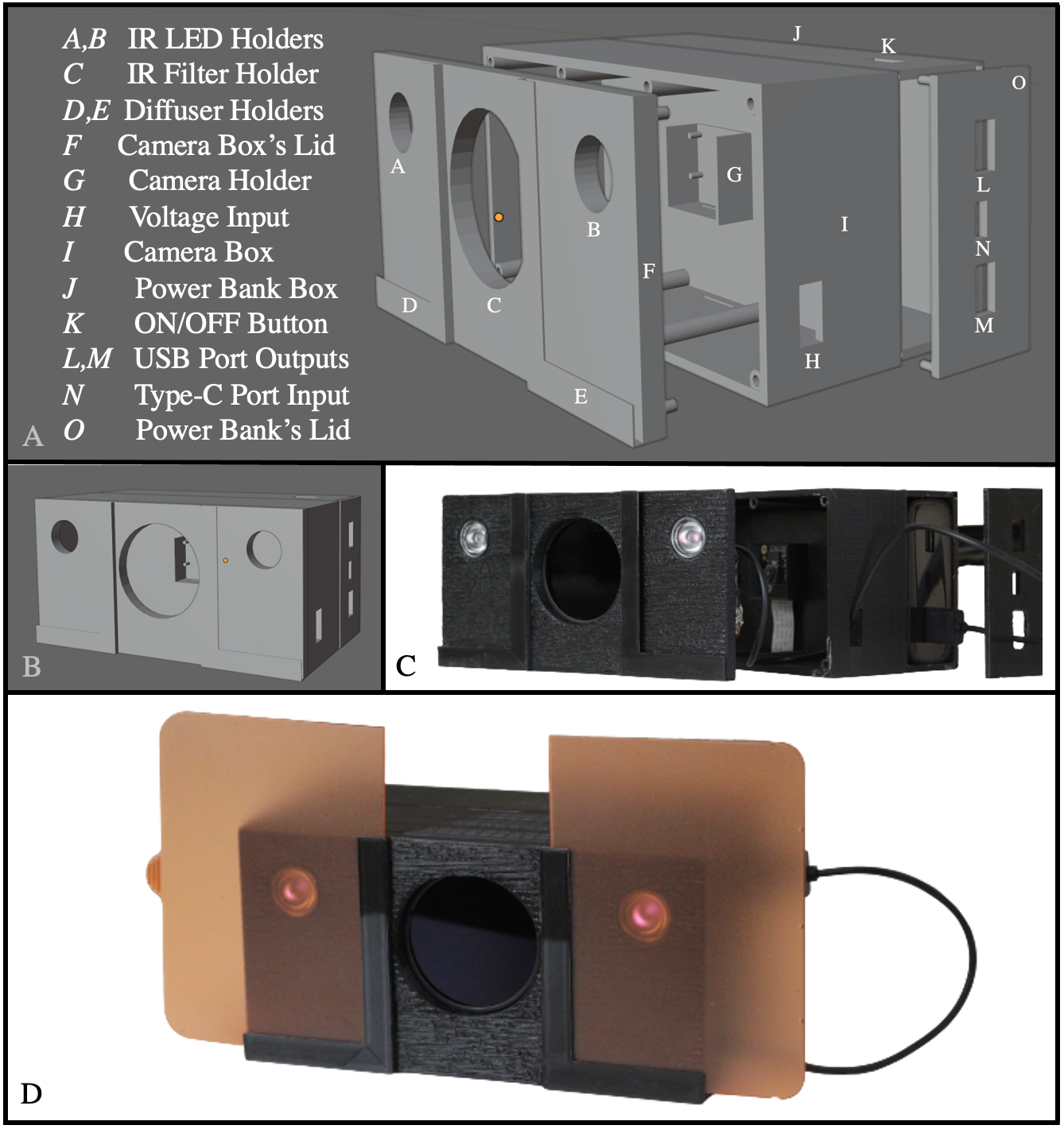}
\caption{Device design for NIR imaging system. (A) 3D modeling of the device. Device modeling was drawn using free and open-source software Blender 2.83. (B) Model of the device. (C) The printout of the model using 3D printer with whole electronics and optical components. (D) Complete powered device with diffusers.}
    \label{fig:fig3}
\end{figure}

The device was modeled using Blender which is a free and open-source 3D creation suite. The 3D model consists of a power bank box, a camera box and their lids. The device was designed to include all the electrical and optical components All components are embedded in the camera box and its lid. A box for the power bank is attached to the back of the camera box. Detachable lids are designed for both boxes to be used when electrical and optical components or power bank need to be changed. The printout of our model was acquired using a 3D printer and all components were placed. The NIR LEDs are powered by the USB cable from the power bank, they are also adjusted to the best brightness with the resistors inside the box. The modeling and printout of the device are shown in Figure \ref{fig:fig3}.

\begin{figure}[htbp] \centering
\includegraphics[width=\linewidth]{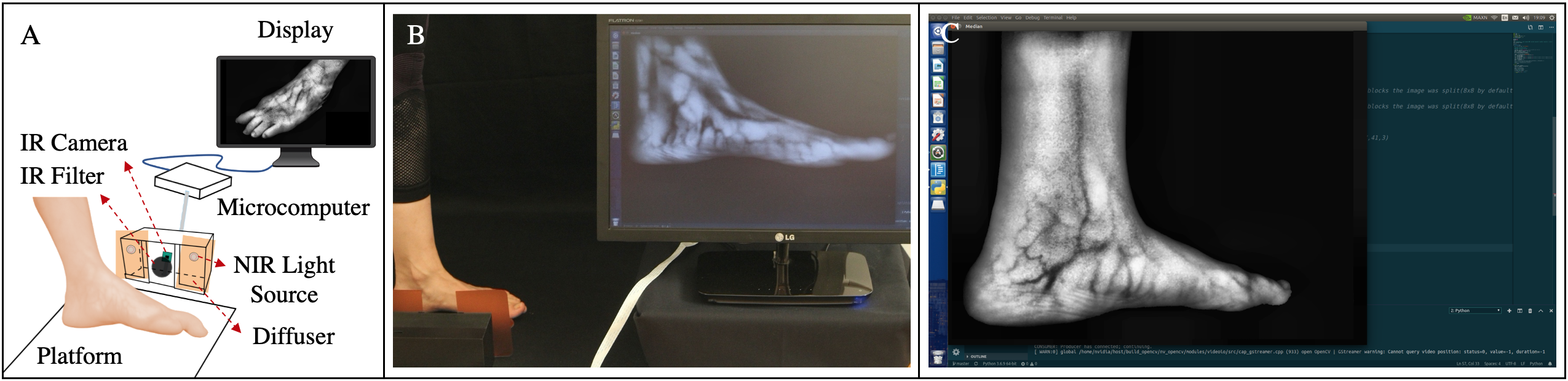}
\caption{Operating structure of the system. (A) Illustration of the hardware system and a real result with the process applied on the screen. (B) Real-time image acquisition from one of the volunteers. (C) A display screenshot of the real-time video processing result.}
    \label{fig:fig4}
\end{figure}

The cost of the system is approximately 160 U.S. Dollars. The whole system while performing real-time processing is shown in the Figure \ref{fig:fig4}. The superficial vein images taken with the device can be seen clearly after background removal and required image processing steps realized in real-time.

\subsection{Software Implementation}

In order to capture good quality vascular images, developing a robust image processing algorithm is as important as the environmental conditions of the system. The images captured using the microcomputer have been processed with OpenCV library on the Python programming language. The process includes grayscale conversion, contrast–limited adaptive histogram equalization (CLAHE), and median filtering. In addition, Frangi filtering was implemented to convert acquired images into tube-like structures. 

Grayscale conversion converts image data in RGB images to a smaller size, simplifying the algorithm and reducing computational requirements. Because a grayscale image is one in which only colors are shades of gray, and less information needs to be provided for each pixel. It is important not to lose the properties of the image during this process. An algorithm has been implemented to preserve the contrast, sharpness, shadow and structure of images \cite{Saravanan2010}. 

\begin{equation}
\label{eqn:eq1}
Y = (0.299 * R) + (0.587 * G) + (0.114 * B)
\end{equation}
\begin{equation}
\label{eqn:eq2}
    U = (B - Y) * 0.565,\,\,\,\,\,V = (R - Y) * 0.713
\end{equation}
\begin{equation}
\label{eqn:eq3}
	UV = U+V
\end{equation}
\begin{equation}
\label{eqn:eq4}
    R_1=R*0.299,\,\,\,\,\,R_2=R*0.587,\,\,\,\,\,R_3=R*0.114
\end{equation}
\begin{equation}
\label{eqn:eq5}
    G_1=G*0.299,\,\,\,\,\,G_2=G*0.587,\,\,\,\,\,G_3=G*0.114
\end{equation}
\begin{equation}
\label{eqn:eq6}
    B_1=B*0.299,\,\,\,\,\,B_2=G*0.587,\,\,\,\,\,B_3=G*0.114
\end{equation}
\begin{equation}
\label{eqn:eq7}
R_4=\frac{R_1+R_2+R_3}{3},\,\,\,\,\,G_4=\frac{G_1+G_2+G_3}{3},\,\,\,\,\,B_4=\frac{B_1+B_2+B_3}{3}
\end{equation}
\begin{equation}
\label{eqn:eq8}
I_1=\frac{R_4+G_4+B_4+UV}{4}
\end{equation}

Initially, ${YUV}$ values expressing luminance of the color and chrominance are calculated from RBG components (Eq. \ref{eqn:eq1} and Eq. \ref{eqn:eq2}). ${Y}$ is the one luminance component that means physical linear-space brightness. The ${UV}$ value represents the total chrominance, ${U}$ is the blue projection and ${V}$ is the red projection (Eq. \ref{eqn:eq3}). The RGB values are then approximated using the RGB components (Eq. \ref{eqn:eq4}–Eq. \ref{eqn:eq7}). The average of these calculated RGB values and the ${UV}$ value gives the ${I_1}$, that is, the resulting gray color image (Eq. \ref{eqn:eq8}). 

\begin{figure}[htbp] \centering
\includegraphics[width=0.75\textwidth]{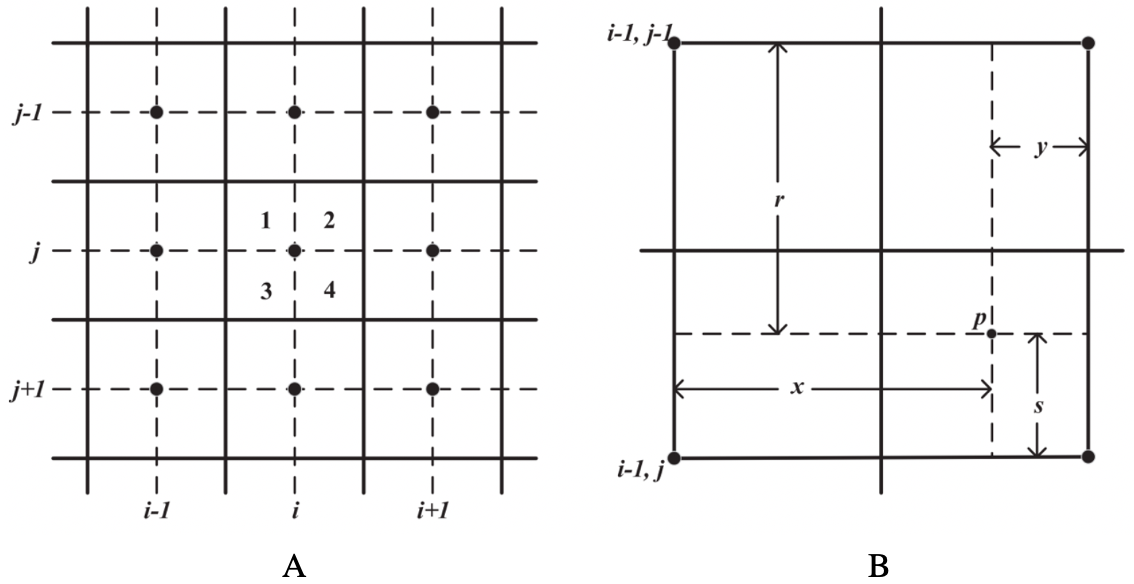}
\caption{The neighborhood structure of inner regions in the CLAHE. (A) A given IR region with its all-neighboring regions. (B) The first quarter of (i,j) region and its relations with the closest four regions \cite{Reza2004}}.
    \label{fig:fig5}
\end{figure}

Image quality is a combination of many factors. Some of those: contrast, blur, lighting, lens flare, noise, and distortion. CLAHE has been used to increase the regional contrast of images in medical imaging \cite{Reza2004,Gupta2018}. In this method, the image is divided into approximately sixty-four identical equal frames and the histogram of each frame is calculated.  Regions consist of three groups; the first group is called the corner regions (CR) in the four corners; the second group, the border regions (BR) in the twenty-four border regions; the remaining thirty-six inner parts are called the inner regions (IR). The clip limit is determined according to the contrast window width and all the regions, each consisting of four quadrants, are recalculated. The number of neighbors of all the parts in the IR group is the same as shown in Figure \ref{fig:fig5}. The new values of the pixels in the IR region are given in Eq.\ref{eqn:eq9}.

\begin{equation}
\label{eqn:eq9}
\begin{array}{cc}
\rho = \frac{s}{s+r}*
    \bigg{(}
    {
    \frac{y}{x+y}*f_{i-1,j-1}\big{(}{\rho_{old}}\big{)}
    +\frac{x}{x+y}*f_{i,j-1}\big{(}{\rho_{old}}\big{)}
    }
    \bigg{)}\\
   +\frac{r}{r+s}
*\bigg{(}
    {
    \frac{y}{x+y}*f_{i-1,j}\big{(}{\rho_{old}}\big{)}
    +\frac{x}{x+y}*f_{i,j}\big{(}{\rho_{old}}\big{)}
    }
\bigg{)}
\end{array}
\end{equation}

The pixel calculation in the second and fourth quadrants of the BR group is the same as the pixel calculation in the IR region. Calculation of new pixel values in the first and third quartiles Eq. \ref{eqn:eq10} is also given.

\begin{equation}
\label{eqn:eq10}
    \begin{array}{cc}
         &  \\\rho_{new} = \frac{s}{s+r}*f_{i,j-1}\big{(}\rho_{old}\big{)}
         +\frac{r}{r+s}*f_{i,j}\big{(}\rho_{old}\big{)}
         & 
    \end{array}
\end{equation}

\begin{figure}[htbp] \centering
\includegraphics[width=0.3\linewidth]{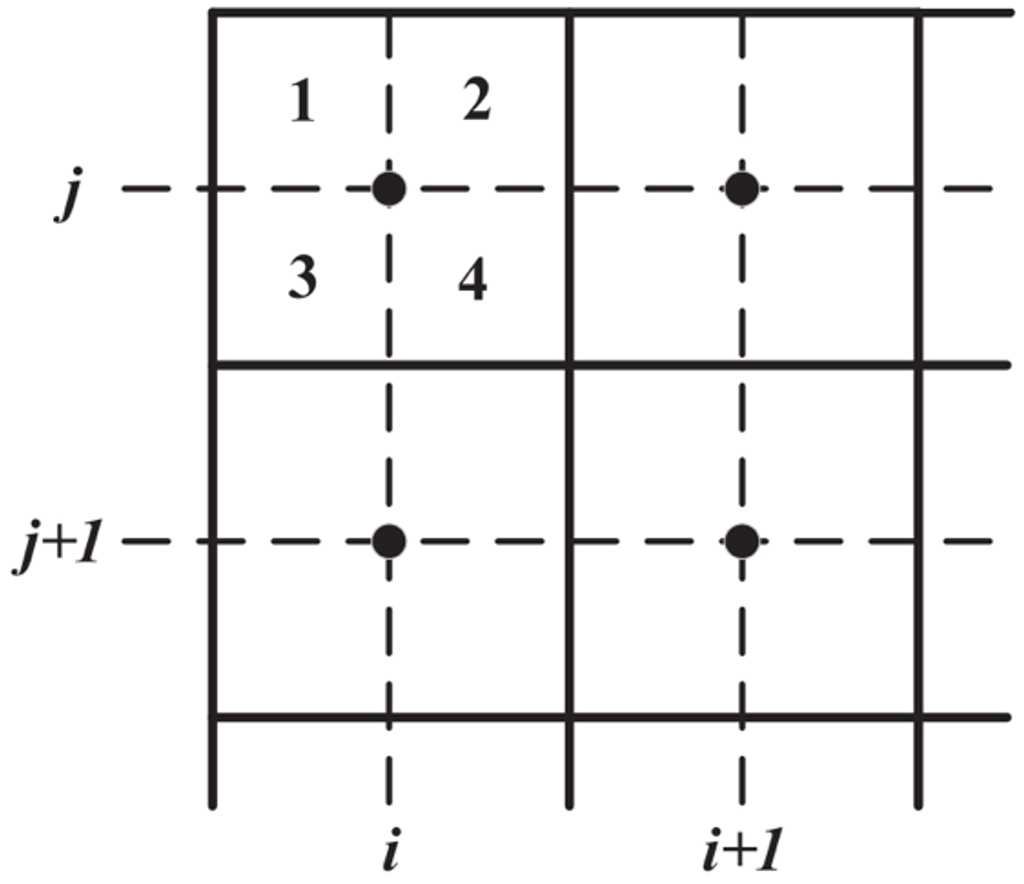}
\caption{The neighborhood structure of a corner region in the CLAHE \cite{Reza2004}.}
    \label{fig:fig6}
\end{figure}

There are three different cases for calculating new pixel values for the CR region. The pixel calculation in the first and third quadrants in the BR region is used for the new pixel values of the second and third quadrants of the CR region. For the pixel values in the fourth quadrant, the pixel value calculation in the IR group is used. The first quadrant of the CR region, as shown in Figure \ref{fig:fig6}, remains the same, as it does not come into contact with other regions (Eq. \ref{eqn:eq11}). 

\begin{equation}
\label{eqn:eq11}
    \rho_{new}=f_{i,j}\big{(}{\rho_{old}}\big{)}
\end{equation}

The median filter is a non-linear filter used to reduce noise in images without interfering with edges and lines. In this method, each pixel of the image is replaced by the median of the pixels in the neighboring regions. The operation is given in Eq. \ref{eqn:eq12} where $\omega$ represents the neighborhood of a pixel with location at $[m,n]$. 

\begin{equation}
\label{eqn:eq12}
    y[m,n]=median\{x[i,j],(i,j)\epsilon\omega\}
\end{equation}

The Frangi filter is a filter to visualize the vascular structure in a tube-like structure and is used in systems such as digital subtraction angiography (DSA), X-ray rotational angiography, computed tomography angiography (CTA), and magnetic resonance angiography (MRA). Hessian matrix eigenvalue analysis is used during the process of vascular structures. By calculating the Gaussian second-order derivative, the Hessian matrix in the point $x$ at scale $\sigma$, $H_{\sigma}\big{(}x\big{)}$ is calculated as in Eq. \ref{eqn:eq13}.

\begin{equation}
\label{eqn:eq13}
    H_{\sigma}\big{(}I,x\big{)}=\frac{\partial^2I_{\sigma}}{\partial{x^2}}=I\big{(}x\big{)}*\frac{\partial^2G_{\sigma}\big{(}x\big{)}}{\partial{x^2}}
\end{equation}

where $I$ is the image and $G_\sigma$ is the gaussian function with the standard deviation $\sigma$. The separation of the second-order structure of the image provides the extraction of eigenvalues $(|\lambda_1 |\leq|{\lambda_2|}\leq{|\lambda_3|})$ and Hessian analysis has an intuitive reason for vascular detection. A vesselness function, $V^{\sigma}_F\big{(}x\big{)}$, was designed to measure the similarity of vascular structures of different sizes to an ideal tube by Frangi $\emph{et al.}$ \cite{Frangi1998} (Eq. \ref{eqn:eq14}).

$V^{\sigma}_F\big{(}x\big{)}$
\begin{equation}
\label{eqn:eq14}
    =\left\{ \begin{array}{rcl} 
    0 & \mbox{if} & \lambda_2 > 0\,\,\,or\,\,\,\lambda_3 > 0 \\ \bigg{(}{1-exp\bigg{(}\frac{R^2_A}{2\alpha^2}\bigg{)}}\bigg{)}exp\bigg{(}\frac{R^2_B}{2\beta^2}\bigg{)}\bigg{(}{1-exp\bigg{(}\frac{\delta^2}{2c^2}}\bigg{)} & else
\end{array}\right.
\end{equation}

Here $\alpha$, $\beta$, and $c$ are parameters that control the sensitivity of the filter and measures dissimilarity that distinguish between tube-like and plate-like structures $(R_A)$ (Eq. \ref{eqn:eq15}), blob-like $(R_B)$ (Eq. \ref{eqn:eq16}) and background $(S)$ (Eq. \ref{eqn:eq17}):

\begin{equation}
\label{eqn:eq15}
   R_A = \frac{|\lambda_2|}{|\lambda_3|}
\end{equation}

\begin{equation}
\label{eqn:eq16}
   R_B = \frac{|\lambda_1|}{\sqrt{|\lambda_2\lambda_3|}} 
\end{equation}

\begin{equation}
\label{eqn:eq17}
    \delta = \sqrt{\lambda^2_1+\lambda^2_2+\lambda^2_3}
\end{equation}

Based on the filter responses depending on the integrals of different scales at the end of the operations, vesselness function is estimated through the maximum response (Eq. \ref{eqn:eq18}).

\begin{equation}
\label{eqn:eq18}
   V_F\big{(}x\big{)} =   \max_{\sigma_{min} \leq \sigma \leq \sigma_{max}} V^{\sigma}_F\big{(}x\big{)}
\end{equation}

\section{Results and Discussions}

In this study, the designed IR imaging device is optimized in terms of software and hardware. Electronic components and external environmental conditions were adjusted to take the best possible images. Then, image processing methods were applied to increase the image quality to get clear vascular structures. After system development, the vascular images were taken from diabetic and non-diabetic volunteers in order to test the performance of the device by observing superficial vascular structures. 

\subsection{Adjusting the Lighting}

\begin{figure}[htbp] \centering
\includegraphics[width=0.75\linewidth]{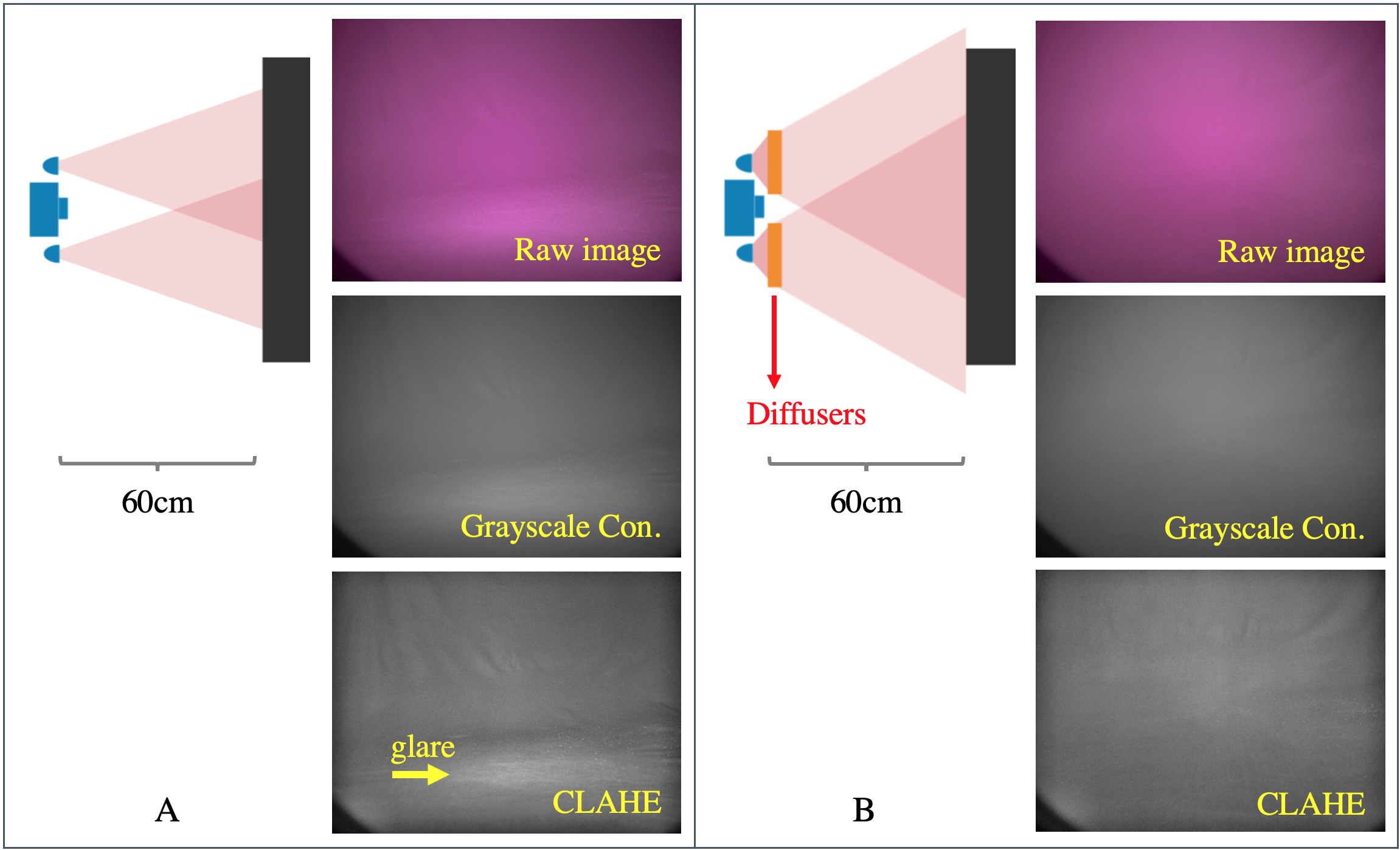}
\caption{The effect of diffusers on the images with a top view of the animation of the imaging system. 2 LEDs and camera are aligned at the same axes to select the best condition during the capturing process with less shadows and glares with and without diffusers. (A) Images captured without diffusers. The glare (yellow arrow) that occurs after the process is clearly visible. (B) Images with the same process applied with diffusers where glare effect is eliminated.}
    \label{fig:fig7}
\end{figure}

Shadow and glare negatively affect blood vessel image analysis. To improve the image quality, shadows and glares must be minimized. In the first section of the adjusting the lighting, the effect of diffusers on glare is examined. As can be clearly seen in the images taken on the backdrop with and without diffusers, there is a reduction in glare in images with diffusers. In Figure \ref{fig:fig7}, daylight images are taken with 2 NIR LEDs to understand the effect of diffusers on image quality. LEDs and camera are aligned at the same axes to select the best condition during the capturing process with less shadows and glares. It was observed that images captured with diffusers have less glare and shadows compared to the images captured without diffusers. 

\begin{figure}[htbp] \centering
\includegraphics[width=0.75\linewidth]{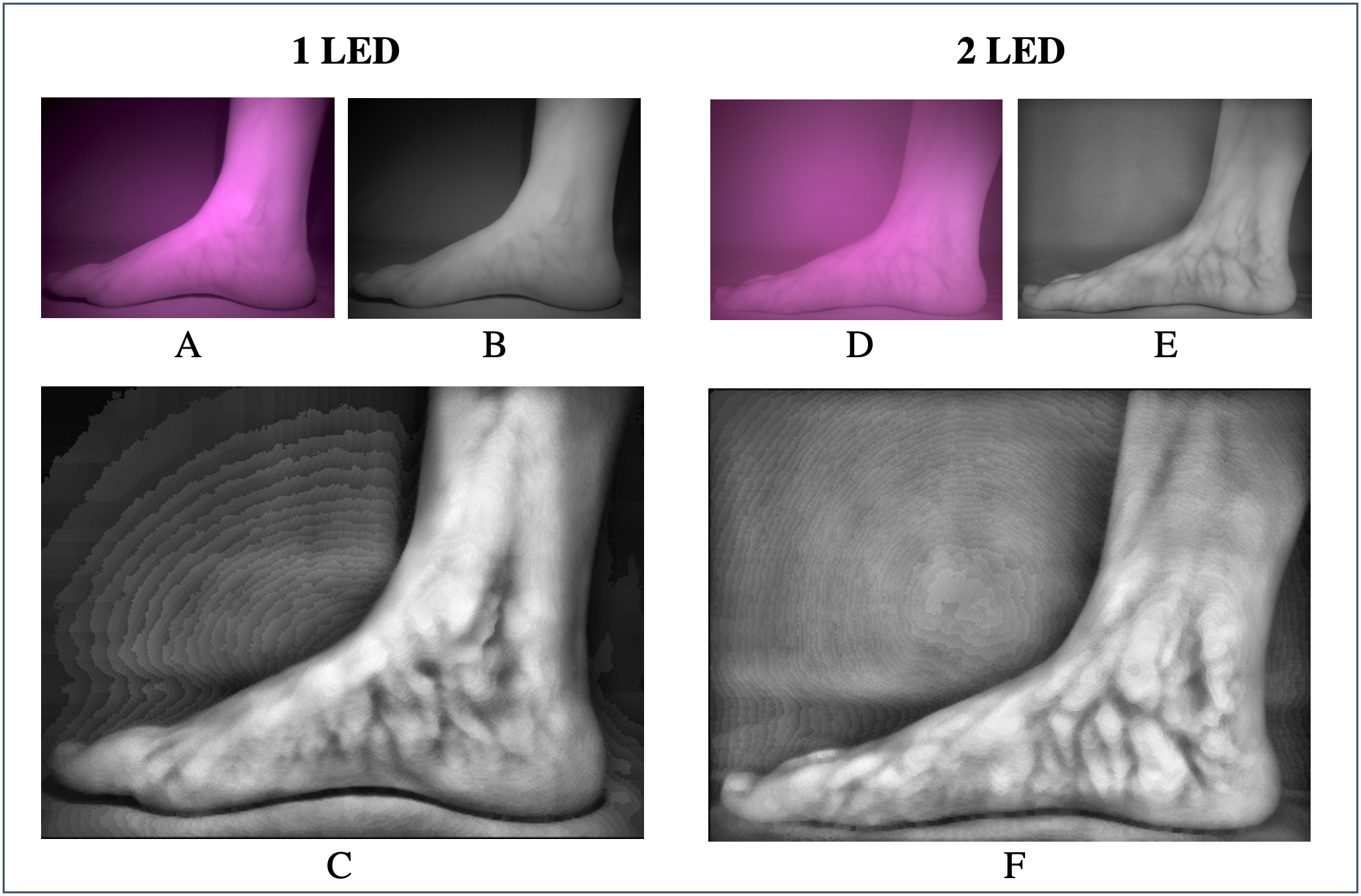}
\caption{The effect of NIR light intensity on the images. The process on the images captured using 1 and 2 NIR LEDs to select the quantity of the used NIR LEDs to achieve minimal data loss. (A, D) Raw images. (B, E) Grayscale conversion. (C, F) CLAHE applied.}
    \label{fig:fig8}
\end{figure}

The second part of improving the illumination is adjusting the amount of light exposed to the imaging area. It has been observed that there is data loss in the images taken when the light is insufficient. On the other hand, shadows and glares increase when the imaging area is over-illuminated. To evaluate the effect of illumination on the imaging area, the images were taken by changing the number of LEDs while the camera and target positions were fixed for the selected LED type. It has been observed that insufficient light emitted by only 1 LED causes data loss, and using 3 or more LEDs causes glare and shadows. As a result, 2 LEDs are selected (Figure \ref{fig:fig8}).

\begin{figure}[htbp] \centering
\includegraphics[width=0.75\linewidth]{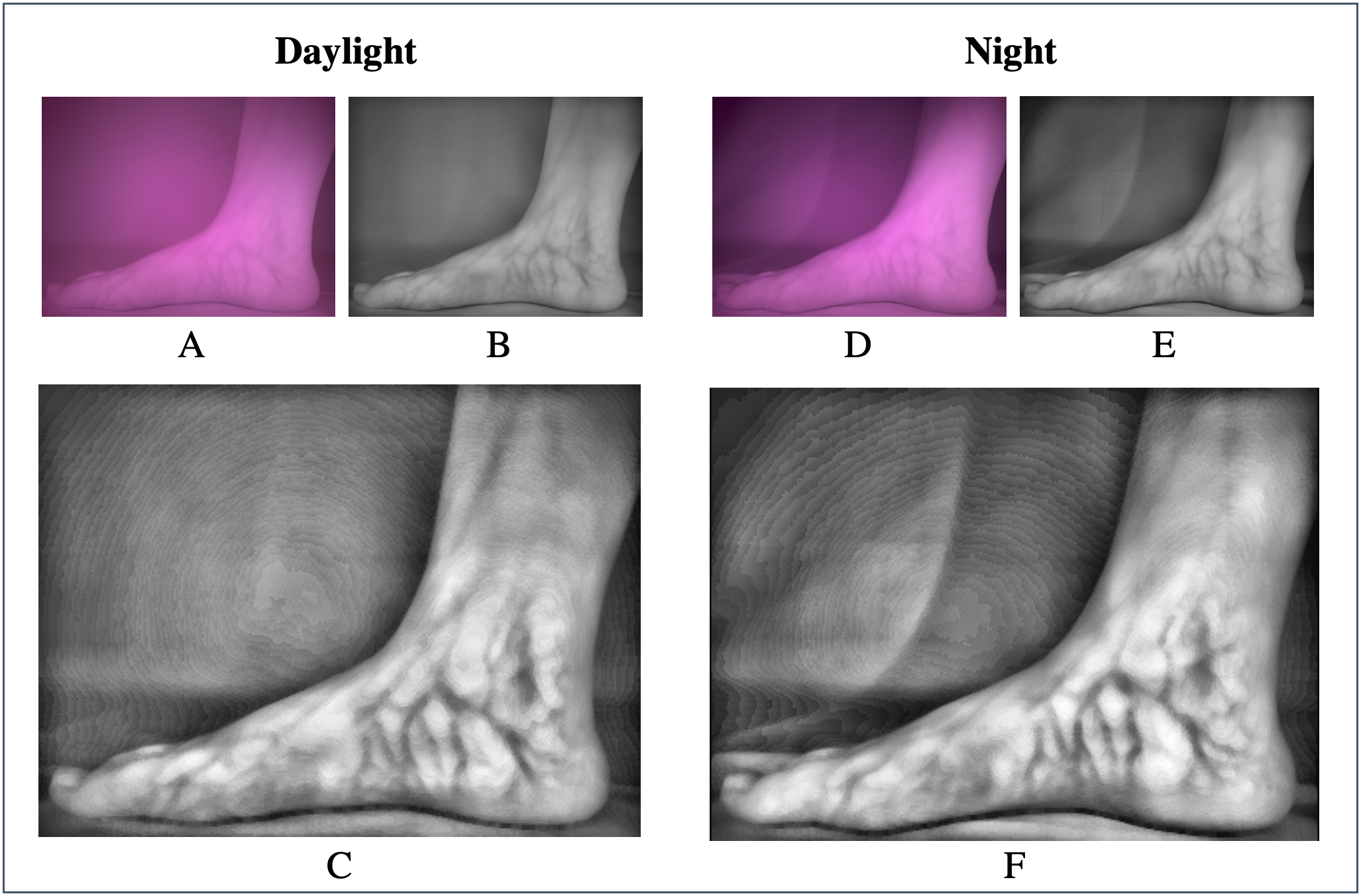}
\caption{Comparison of the images taken during daylight and at night to observe the effect of environment illumination and the quality of the pictures taken. LEDs and camera are aligned at the same axes, night and daylight images captured with 2 NIR LEDs to observe diffusing of the light. (A, D) Raw images. (B, E) Grayscale conversion. (C, F) CLAHE applied.}
    \label{fig:fig9}
\end{figure}

The final part analyzed in adjusting the effectiveness of lighting is the comparison of night versus daylight imaging. Environment was kept completely dark during night imaging. Even though shadows were increased when the images were taken at night, data loss was not observed compared to taking images in daylight. Nevertheless, daylight images give better result due to non-glare (Figure \ref{fig:fig9}). This shows that there is no need to create a private dark environment to take the images.

\subsection{Adjusting the Positions of Target and LEDs}

\begin{figure}[htbp] \centering
\includegraphics[width=0.75\linewidth]{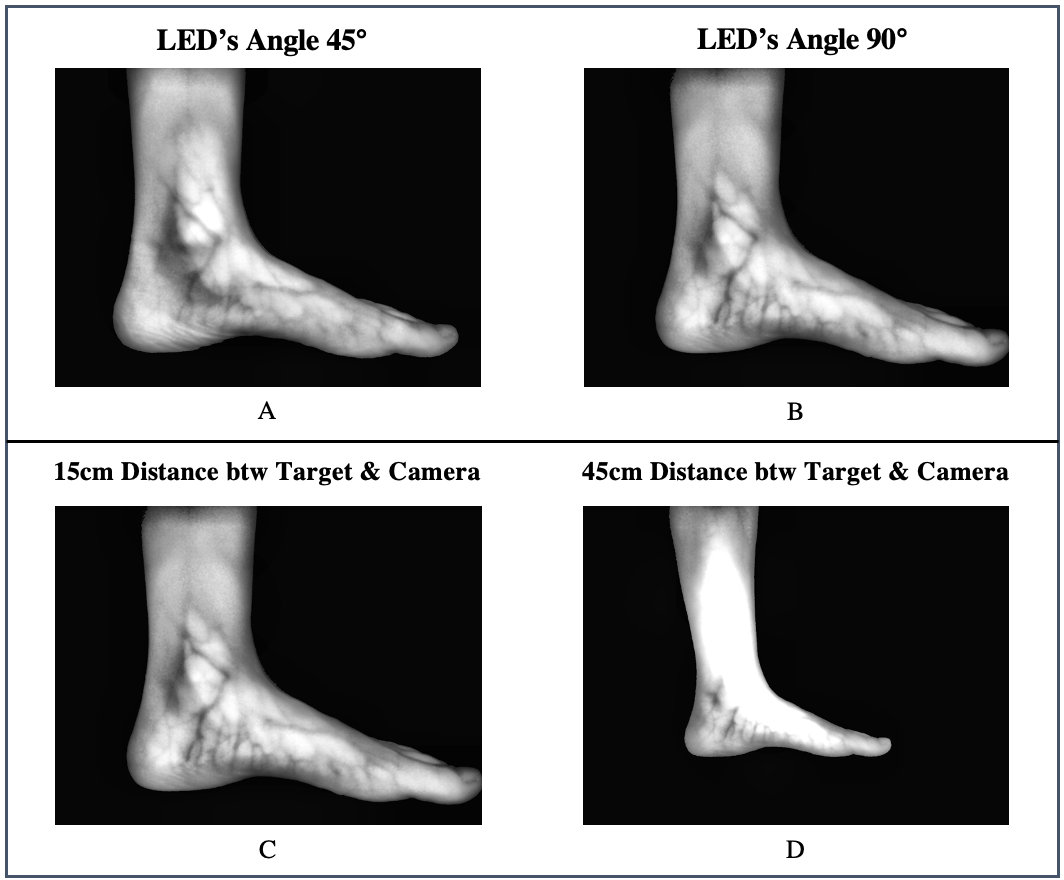}
\caption{Adjusting the lighting of the system. (A, B) Images taken by changing the angles of the LEDs to test for decreasing the glare effect. (C, D) Images captured by changing the distance of the target to the camera to select the optimal distance between the device and the target with less glare and data loss.}
    \label{fig:fig10}
\end{figure}

Environmental conditions of the system along with the location of the camera and target were important to be able to get a good image by reducing the glare while increasing the contrast. It is known that the angle of irradiated light affects the captured images \cite{Kato2012}. The LED angle was adjusted in two different ways as an estimated 45 degrees and 90 degrees perpendicular to the target and evaluations were made on the captured images. It was clearly seen in the results that the vascular structures in the images captured at 90 degrees were more noticeable than those captured at 45 degrees in the Figure \ref{fig:fig10}(A, B). 

The optimum distance of the target to the camera is an important factor to improve the image quality. As the target moves away from the camera, data loss occurs due to the light glare. Result of these tests indicates that the distance between backdrop and device should be 60 cm to acquire good quality images. The target also should be positioned 15 cm away from the device. The images are taken from different distances are shown in the Figure \ref{fig:fig10}(C, D).

\begin{figure}[htbp] \centering
\includegraphics[width=0.75\linewidth]{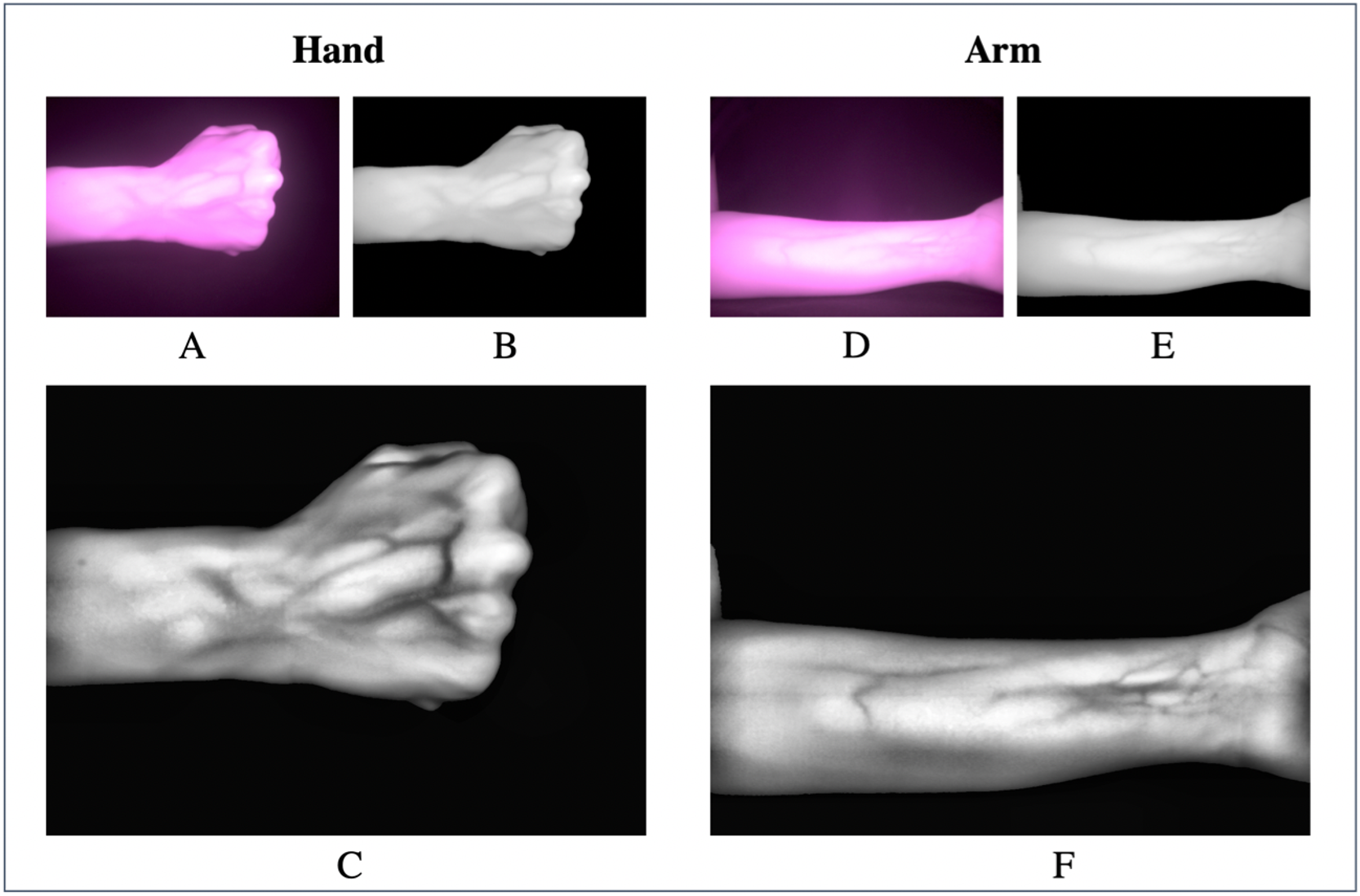}
\caption{Demonstrating the effectiveness of the device for superficial vein images of the arm and hand. (A, D) Raw images. (B, E) Images with removed background. (C, F) CLAHE applied.}
    \label{fig:fig11}
\end{figure}

After the optimum conditions are provided, the hand and arm images were also taken which shows that the device is effective to image the superficial veins in different parts of the body (Figure \ref{fig:fig11}).

\subsection{Real-time Video Processing}

One of the most important features that distinguishes the developed system is the ability to do real-time video processing to show superficial vascular structures. During real-time video processing, first of all, a frame appropriate for the target region is created. Next, the background is removed from the image and then the 3D RGB image is converted to a 2D black and white image with grayscale conversion in order to shorten the processing time and reduce the data. Afterwards, the median filter is applied on the 2D image to remove the noise such as hair. Finally, CLAHE is applied three times with different grid sizes to the 2D image to increase the contrast and sharpen the image. The reason for using different grid sizes is to avoid unwanted noises caused by window edges in overlapping CLAHE’s layer blocks. Selected grid size values are 4x4 for the first layer, 10x10 for the second layer, and 16x16 for the third layer. The whole process is rapid and can be done in real-time using a portable low power microcomputer (NVIDIA Jetson Nano) to show clear superficial vascular structures.

\subsection{Frangi Filtering for tubular visulatizion of the blood vessels}

\begin{figure}[htbp] \centering
\includegraphics[width=0.75\linewidth]{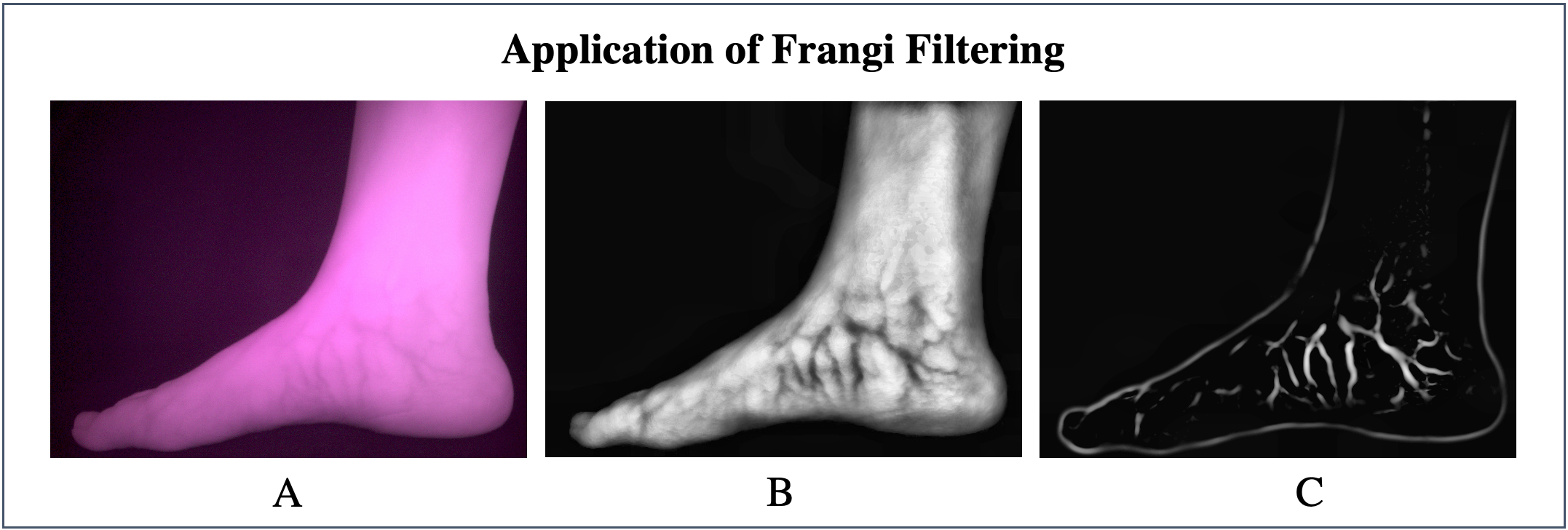}
\caption{Frangi filtering application to the superficial veins of the foot to detect vessel-like or tube-like structures. (A) Raw image with background removed. (B) Raw image with CLAHE applied. (C) Frangi filtering followed by CLAHE.}
    \label{fig:fig12}
\end{figure}

In addition to real-time video processing, the captured vessel images were converted into a tube-like structure using a Frangi filter. The Frangi filter, developed by Frangi $\emph{et al.}$ \cite{Frangi1998}, identifies a specific function by selecting certain eigenvalues to improve blood vessels display. It suppresses non-vascular structures and background noise while increasing the contrast of the blood vessels. The multi-scale second-order local structure of the images is examined and the tubular vascular structure image is revealed. The results of the Frangi filtered images are given in Figure \ref{fig:fig12}. However, this filter is not suitable for real-time processing as it extends the analysis time (approximately 3 minutes) using the Jetson Nano.

\subsection{Analysis of Images Captured from Volunteers}

\begin{figure}[htbp] \centering
\includegraphics[width=0.75\linewidth]{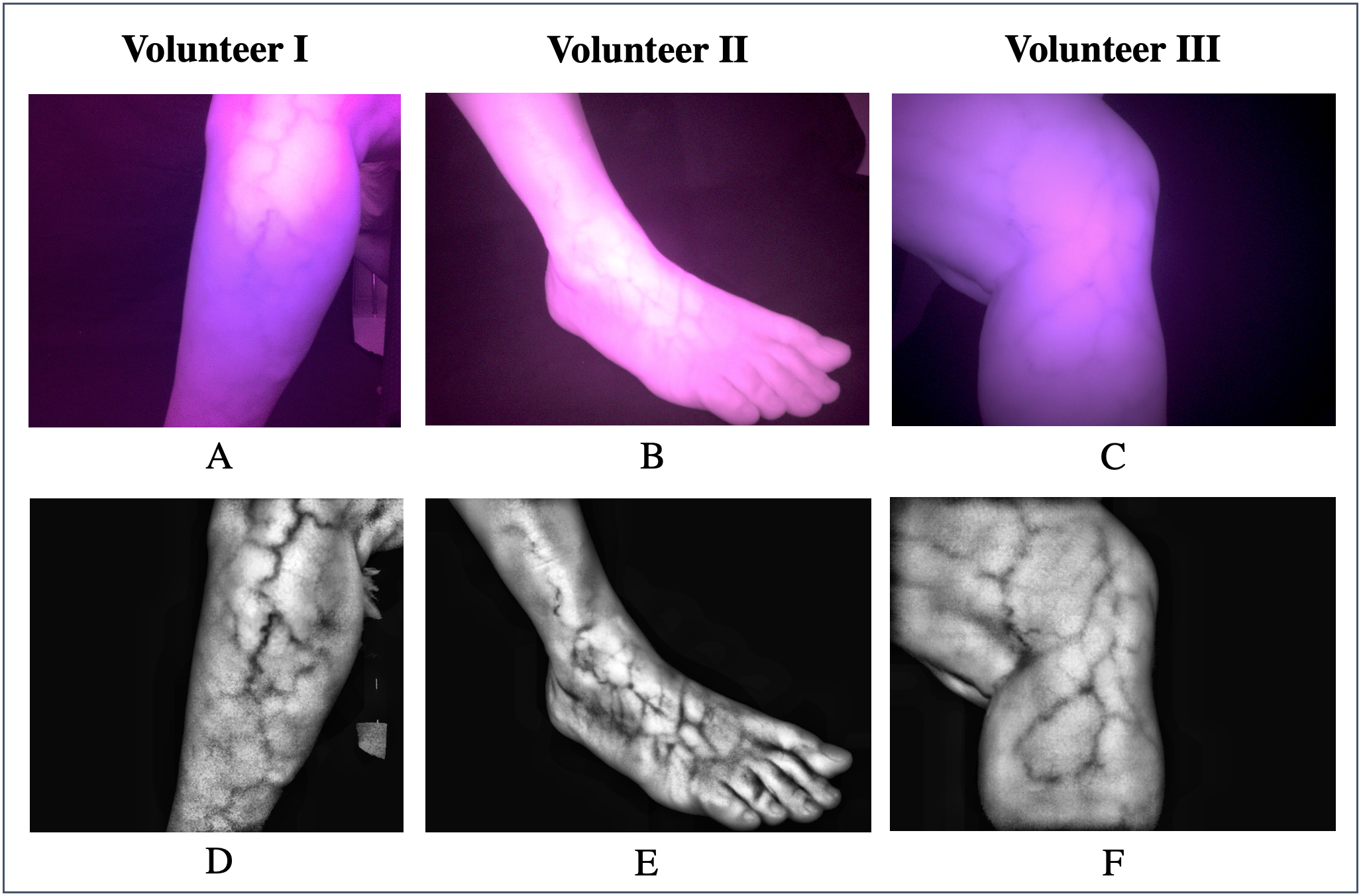}
\caption{Vascular structure images of the lower leg and foot taken from diabetic volunteers (Volunteers I, II, III) , who are expected to have abnormal vascular structure because of deformations caused by high blood sugar in the blood. Raw images and output superficial vein images after all processing are represented for each diabetic volunteer, respectively: (A, D) elderly diabetic person “Volunteer I” with toe amputation due to diabetic foot, (B, E) a middle-aged person (Volunteer II) with diabetes using insulin, (C, F) an elderly volunteer with newly diagnosed diabetes uses only medicine, not insulin (Volunteer III).}
    \label{fig:fig13}
\end{figure}

\begin{figure}[htbp] \centering
\includegraphics[width=0.75\linewidth]{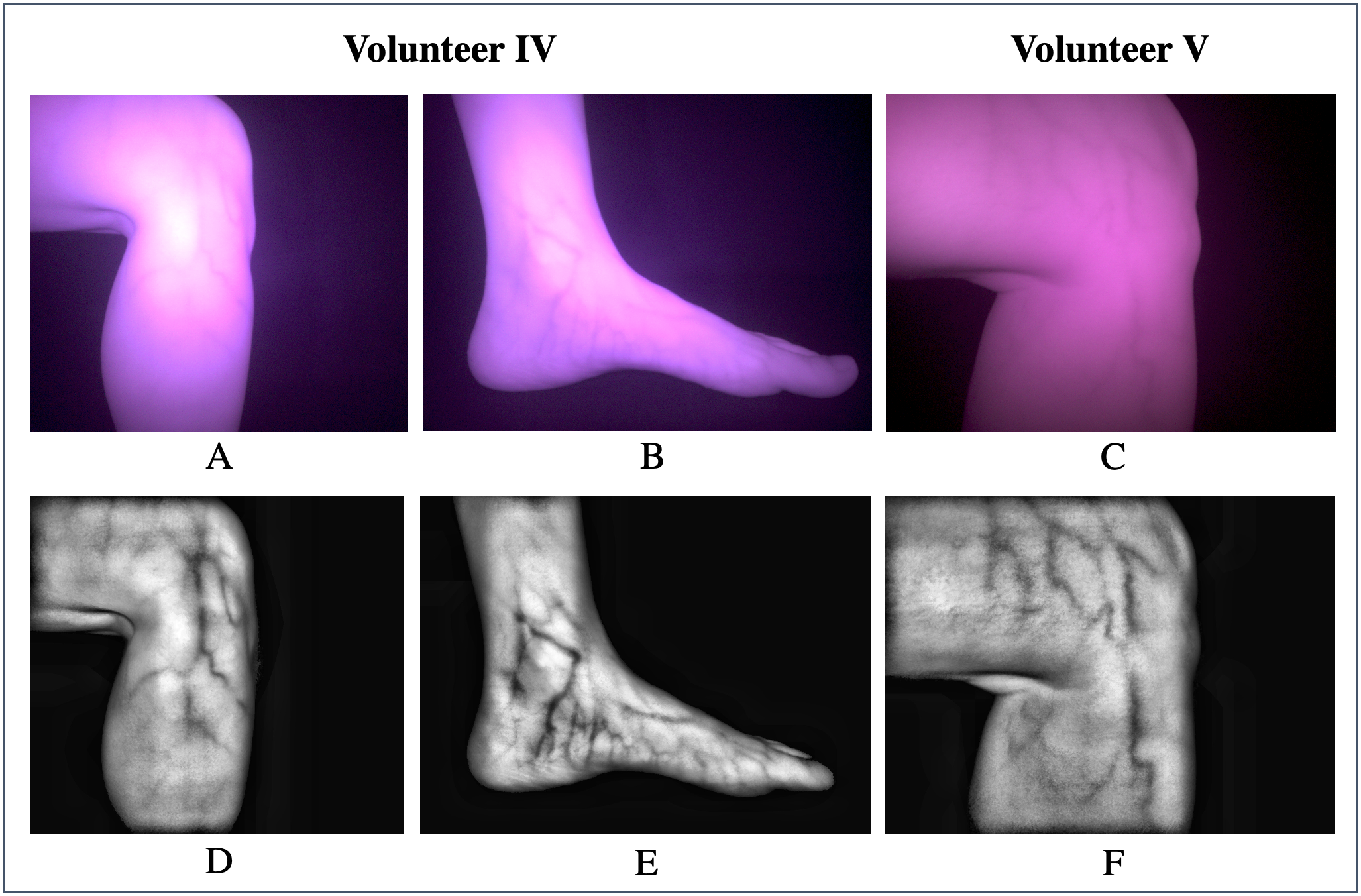}
\caption{Vascular structure images of the lower leg and foot taken from non-diabetic volunteers (Volunteers IV, V)  who are expected to have healthier vascular structure compared to volunteers with diabetes. (A, B, C) Raw images. (D, E, F) Output superficial vein images after all processing are applied. (A, B, D, E) Images belong to Volunteer IV and (C, F) images belong to Volunteer V.}
    \label{fig:fig14}
\end{figure}

Since the fibrinolytic response to venous occlusion is known to be weaker in diabetic patients than in non-diabetics, the volunteers were examined under two classes: diabetic and non-diabetic, to observe superficial vein abnormalities. In this study, five participants volunteered for image analysis. Three of these volunteers have diabetes at different stages, and two of them are non-diabetic volunteers. Volunteer I had a toe amputation due to diabetic foot complications. Volunteer II had diabetes around middle age and must take insulin. Volunteer III has just been diagnosed with diabetes and just using the medication without the need for insulin injections. Volunteer IV and Volunteer V subjects do not have diabetes. 
Foot superficial vein images taken with our device were processed for each volunteer, and the results were compared. While circulatory abnormalities of Volunteers I, II, and III (Figure \ref{fig:fig13}) were observed, significant abnormalities were not seen in Volunteers IV and V (Figure \ref{fig:fig14}). The cause of the abnormalities in the vascular structures might be diabetes or other unrelated diseases to diabetes, such as cardiac dysfunction or varicose vein. In addition, smoking, obesity, unhealthy diet, inactivity, and advanced age reasons may also be the cause of the abnormalities. It is known that slightly tortuous and twisted arteries and veins are seen in humans and animals, and this is a common anomaly without symptoms \cite{weibel1965tortuosity,schep2001magnetic,helisch2003arteriogenesis}. However, observed images show that mild tortuosity in superficial veins is excessive in diabetic volunteers, and severe tortuosity can lead to serious symptoms. Here we again need to emphasize that this is an engineering study and does not have an intention to diagnose any diseases. We are also not claiming that we developed a diagnostic device alternative to other methods. However, the developed system here might assist healthcare personnel in early diagnosis and treatment follow-up of treatment processes for superficial vascular structures such as sclerotherapy and may create new opportunities. In addition, the developed system can be used for other medical and security applications such as blood vein detection during injection or a bio-metric validation system with palm trail and finger veins. The system has advantages in terms of being low-cost, user-friendly, and non-contact. On the other hand, examining the images over a certain time period might become more meaningful in terms of understanding the course of the deterioration over time. In order to make a clear inference about the clinical usage of the method, the results should be compared to other imaging methods such as angiography and interpreted by a medical specialist.

\FloatBarrier 
\section{Conclusion}
\label{sec:con}
Abnormalities in the blood circulation cause many negative conditions that affect organs and other systems in the human body, such as delayed healing of wounds and ulcers. Early diagnosis of abnormalities in the circulatory system is vital in terms of timely treatment or raising awareness of the patient’s health.
In this study, a low-cost and portable microcomputer-based tool has been developed as a real-time NIR superficial vascular imaging device. The cost of the system is approximately 160 U.S. Dollars. The optimum device and external environmental conditions such as the effect of diffusers, IR illumination intensity, and the distance between the device to target were tested and established to produce the clearest superficial vein images. It was observed that the optimum distance between device and target is 15 cm and the diffusers significantly reduced the glare effect in the images. Two 1 W high power 850 nm NIR LEDs were selected as the illumination intensity, where glare and shadows have occurred on the images with more or less lighting. All of the image and video analysis was accomplished on a portable low-power embedded microcomputer, NVIDIA Jetson Nano, using OpenCV. 
The developed device was tested on three diabetic and two non-diabetic volunteers. Vascular structural changes that occur due to high sugar accumulation in the blood circulation are some of the complications caused by diabetes. These changes make it easier to observe vascular abnormalities in people with diabetes. When the images from diabetic versus non-diabetic volunteers were compared, abnormalities in the superficial vascular structures of the people with diabetes were observed, as expected. As a result, tortuosity was observed successfully in the superficial vascular structures, where the results need to be interpreted by the medical experts in the field with further investigations to understand the underlying reasons. Additionally, as a future work, it is recommended to realize a comparative study where the images taken from the same volunteers with the designed device and the angiography are compared and interpreted by medical experts to better understand the accuracy of the device. Although this study is an engineering study and does not have an intention to diagnose any diseases, the developed system here might assist healthcare personnel in early diagnosis and treatment follow-up for vascular structures and may enable further opportunities.

\section*{Acknowledgments}
We would like to thank Assoc. Prof. Dr. Sevket Gumustekin for laboratory resources during 3D modeling of the device. We also would like to thank you to Izmir Institute of Technology for their support in this project by Scientific Research Projects Coordination Unit (BAP) 2020IYTE0112.

\section*{Declarations}

\textbf{Conflict of interest} The authors have no relevant financial or non-financial interests to disclose. 

\begin{justify}
\justifying
\textbf{Ethical approval} All the imaging procedures involving human participants were conducted by the approval of Ethical Committee of Izmir Institute of Technology (No: 02.10.2020-E.21015). Informed consent was obtained from all individual participants in this study. 
\end{justify}

\bibliographystyle{unsrt}  
\bibliography{references.bib}

\end{document}